\documentclass[sigconf]{acmart}

\copyrightyear{2020}
\acmYear{2020}
\setcopyright{acmcopyright}
\acmConference[KDD '20] {26th ACM SIGKDD Conference on Knowledge Discovery and Data Mining}{August 23--27, 2020}{Virtual Event, USA}
\acmBooktitle{26th ACM SIGKDD Conference on Knowledge Discovery and Data Mining (KDD '20), August 23--27, 2020, Virtual Event, USA}
\acmPrice{15.00}
\acmDOI{10.1145/XXXXXX.XXXXXX}
\acmISBN{978-1-4503-7998-4/20/08}

\usepackage{graphicx}
\usepackage{balance}  %

\usepackage{latexsym}
\usepackage{amsfonts}
\usepackage{amsmath}
\usepackage{amssymb}
\usepackage{color}
\usepackage{epsfig}
\usepackage{xspace}
\usepackage{subfigure}
\usepackage{rotating}
\usepackage{pbox}
\usepackage{caption}
\usepackage{hyperref}
\usepackage{breakurl}
\usepackage{epsfig}
\usepackage{multirow}
\usepackage{url}
\usepackage{enumitem}

\makeatletter
\DeclareRobustCommand*\cal{\@fontswitch\relax\mathcal}
\makeatother

\newcommand{\myhrule}{\rule[.5pt]{\hsize}{.5pt}}

\newcommand{\eat}[1]{}

\newcommand{\sstab}{\rule{0pt}{8pt}\\[-2.4ex]}

\newcommand{\bi}{\begin{itemize}}
\newcommand{\ei}{\end{itemize}}
        {\end{itemize}\vspace{-0.5ex}}

\newcommand{\mat}[2]{{\begin{tabbing}\hspace{#1}\=\+\kill #2\end{tabbing}}}

\newcommand{\be}{\begin{enumerate}}
\newcommand{\ee}{\end{enumerate}}
\newcommand{\beqn}{\begin{eqnarray*}}
\newcommand{\eeqn}{\end{eqnarray*}}

\newcommand{\stitle}[1]{\vspace{.5ex}\noindent{\bf #1}}
\newcommand{\eetitle}[1]{\vspace{0.8ex}\noindent{\underline{\em #1}}}

\newcommand{\ie}{\emph{i.e.,}\xspace}
\newcommand{\eg}{\emph{e.g.,}\xspace}

\newcommand{\IN}{\mbox{{\bf in}}\ }
\newcommand{\If}{\mbox{\bf if}\ }

\newcommand{\Then}{\mbox{\bf then}\ }

\newcommand{\Else}{\mbox{\bf else}\ }

\newcommand{\Do}{\mbox{\bf do}\ }

\newcommand{\ForEach}{\mbox{\bf for each}\ }

\renewcommand{\And}{\mbox{\bf and}\ }

\newcommand{\Return}{\mbox{\bf return}\ }

\newcommand{\kw}[1]{{\ensuremath {\mathsf{#1}}}\xspace}

\newcounter{ccc}
\newcommand{\bcc}{\setcounter{ccc}{1}\theccc.}
\newcommand{\icc}{\addtocounter{ccc}{1}\theccc.}

\newcommand{\eop}{\hspace*{\fill}\mbox{$\Box$}}     %
\newcounter{definition}%
\renewcommand{\thedefinition}{\arabic{definition}}

\newcounter{example}%
\renewcommand{\theexample}{\arabic{example}}
\newenvironment{example}{
        \vspace{0.8ex}
        \refstepcounter{example}
        {\noindent\bf Example \theexample:}}{
        \eop\vspace{0.8ex}}

\newcounter{prop}%
\renewcommand{\theprop}{\arabic{prop}}

\renewcommand{\eop}{\hspace*{\fill}\mbox{$\Box$}}     %

\renewcommand{\texttt}[1]{{\small\textsf{#1}}}

\definecolor{gray}{rgb}{0.5,0.5,0.5}

\renewcommand{\S}{{\cal S}}

\newcommand{\eval}{\kw{{Match}}}

\newcommand{\LRs}{\kw{{PRs}}}
\newcommand{\LR}{\kw{{PR}}}

\newcommand{\genWIR}{\kw{GenWIR}}
\newcommand{\genLR}{\kw{GenPR}}
\newcommand{\kb}{\kw{KB}}

\newcommand{\annotAlg}{\kw{Annotation}}
\newcommand{\tfm}{features}
\newcommand{\tgt}{labels}

\newcommand{\hp}{hyperparameter}

\newcommand{\guideEval}{\kw{GuideEval}}

\newcommand{\ntbk}{\kw{NTBK}}
\newcommand{\kaggle}{\kw{Kaggle}}

\usepackage{epstopdf}
\setcopyright{acmcopyright}
\copyrightyear{2020}
\acmYear{2020}
\acmDOI{10.1145/1122445.1122456}

\begin{document}
\fancyhead{}

\title{Vamsa: Automated Provenance Tracking in \\ Data Science Scripts}

\author{Mohammad Hossein Namaki}
\email{m.namaki@wsu.edu}
\affiliation{%
  \institution{Washington State University}
}

\author{Avrilia Floratou}
\email{avflor@microsoft.com}
\affiliation{%
  \institution{Microsoft}
}

\author{Fotis Psallidas}
\email{fopsalli@microsoft.com}
\affiliation{%
  \institution{Microsoft}
}

\author{Subru Krishnan}
\email{subru@microsoft.com}
\affiliation{%
  \institution{Microsoft}
}

\author{Ashvin Agrawal}
\email{asagr@microsoft.com}
\affiliation{%
  \institution{Microsoft}
}

\author{Yinghui Wu}
\email{yxw1650@case.edu}
\affiliation{%
  \institution{Case Western Reserve University}
}

\author{Yiwen Zhu}
\email{zhu.yiwen@microsoft.com}
\affiliation{%
  \institution{Microsoft}
}

\author{Markus Weimer}
\email{markus.weimer@microsoft.com}
\affiliation{
 \institution{Microsoft}
}

\begin{abstract}

\eat{Machine learning (ML) which was initially adopted for search ranking and recommendation systems has firmly moved into the realm of core enterprise operations like sales optimization and preventative healthcare. For such ML applications, often
deployed in regulated environments, the standards for user privacy, security, and data governance are substantially higher. This imposes the need for tracking provenance end-to-end, from the data sources used for training ML models to the predictions of the deployed models.}

There has recently been a lot of ongoing research in the areas of fairness, bias and explainability of machine learning (ML) models due to the self-evident or regulatory requirements of various ML applications. We make the following observation: All of these approaches require a robust understanding of the relationship between  ML models and the data used to train them.
In this work, we introduce the \textit{ML provenance tracking} problem: 
the fundamental idea is to automatically track which columns in a dataset have been used to derive the features/labels of an ML model.
We discuss the challenges in capturing such information in the context of Python, the most common language used by data scientists. 

We then present Vamsa, a modular system
that extracts provenance from Python scripts \textit{without requiring any changes to the users' code}. Using $26K$ real data science scripts, we verify the effectiveness of Vamsa in terms of coverage, and performance. We also evaluate Vamsa's accuracy on a smaller subset of manually labeled data. Our analysis shows that Vamsa's precision and recall range from $90.4\%$ to $99.1\%$ and its latency is in the order of milliseconds for average size scripts. Drawing from our experience in deploying ML models in production, we also present an example in which Vamsa helps automatically identify models that are affected by data corruption issues.

\end{abstract}



\maketitle

\section{Introduction}
\label{sec:intro}

 Increasingly, machine learning is used in domains which demand great care and attention to fairness, correctness and reliability~\cite{gebru2018datasheets} such as the financial, medical and manufacturing industries. %
Of course, we are not the first to recognize this trend. In fact, KDD itself has been host to several workshops~\cite{expalinability_workshop}, tutorials~\cite{expalinability_tutorial,fairness_tutorial} and research contributions~\cite{DBLP:conf/kdd/ShuCW0L19, Lee2018WinnersCB} in the areas of fairness, bias and explainability of ML models. This year's call for papers specifically asks for contributions in those areas. Considering these and other works, we make a very simple, and somewhat obvious observation: All these approaches require a robust understanding of the model's provenance: What data was used to train the model?  Where did the training data originate? How was it processed?  These and other questions are usually assumed to have an answer. 

We conducted a survey across 7 Big Data companies to better quantify the need for provenance tracking in the ML space. The participants were responsible for cleaning data, developing, or deploying ML models in production. According to $82\%$ of the participants, tracking provenance between data and  ML models can be useful in multiple scenarios such as: model debugging
($88\%$ of the participants), model sharing ($69\%$), compliance ($56\%$), and fairness ($56\%$). Most of the participants also pointed out that their teams currently spend multiple hours per week trying to identify information related to the data used to train ML models (files, columns used for features, etc.) either manually or using primitive tools.

Projects such as MLflow~\cite{mlflow} and Kubeflow~\cite{kubeflow} make it easy to run \emph{and record} complex ML pipelines. However, these systems do not currently provide  a way to manually or automatically track the relationships between data and ML models. %

\begin{figure}[tb!]
\centering
\centerline{\includegraphics[scale=0.85]{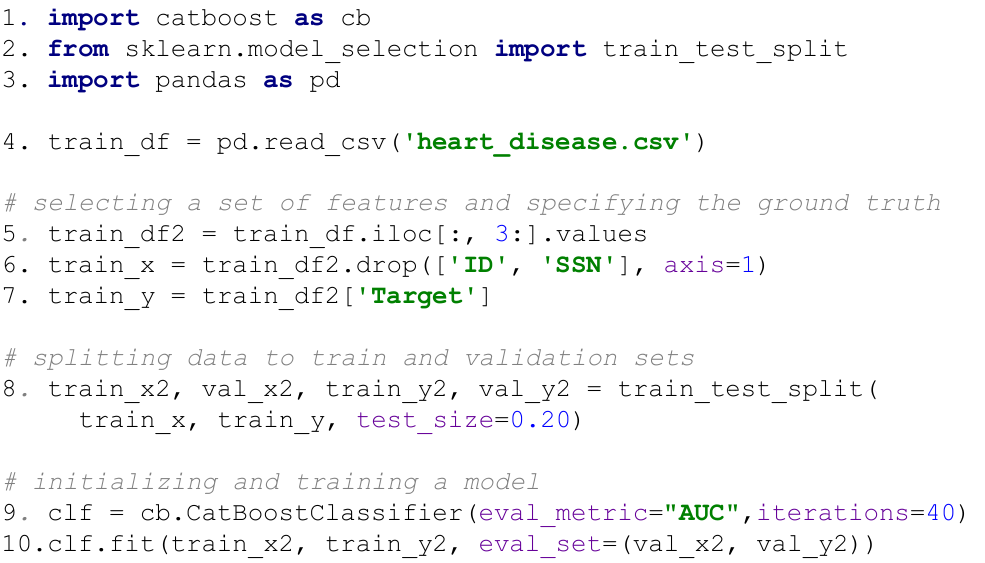}}
\vspace{-1ex}
 \caption{{\small A data science script written in Python}}
\label{fig-motivation-example}
\vspace{-2ex}
\end{figure}

Building upon and going beyond these insights and works, we here present our approach to \emph{automatic} provenance tracking for ML pipelines that seamlessly tracks which columns in a dataset have been used to derive the features and labels of a ML model. Consider the Python script presented in Figure~\ref{fig-motivation-example} that was created in the context of the Kaggle Heart Disease competition~\cite{kaggle_heart_disease}. The script trains a ML model using a patient dataset from a hospital. The model takes as input a set of features such as \texttt{Age} and \texttt{Blood pressure}, and predicts whether a patient might have a heart disease in the future. A practical provenance tracking system should not only detect that this script trains a ML model but also that the model is trained using the \texttt{heart\_disease.csv} dataset and that the columns \texttt{Target} and \texttt{SSN} are not used to derive the model's features. Having this infomation can help detect violations of compliance regulations such as using PII information (e.g., SSN) when training a ML model. %

While this step is obvious in terms of the problem definition, it is far from it in realization: ML pipelines are typically authored in Python, which is beloved for its dynamic typing and extensive meta-programming abilities. The absence of declarative semantics  makes automated provenance tracking particularly challenging. Moreover, data science is an evolving field as exemplified by the growth of newly available frameworks like PyTorch~\cite{pytorch} and popular libraries like scikit-learn~\cite{pedregosa2011scikit} still under active  development.

We argue that a fully automated provenance system for data science scripts to be usable,  should rely on the following important design principles: (1) provide support for unmodified Python scripts (the most common language used by data scientists~\cite{KaggleSurvey}) and (2) be extensible to accommodate new ML frameworks/libraries. In other words, the system must strike a balance between two relatively conflicting objectives: on the one hand, input from the user (e.g., in the form of  logging operations) is not acceptable as it is a time-consuming and potentially error-prone process, on the other hand support for existing and new ML frameworks/libraries is required. %

Towards these goals, we make the following contributions: 
\begin{enumerate}[leftmargin=*,partopsep=1ex,parsep=1ex]

\item We present \textbf{Vamsa}, to the best of our knowledge, the first automated provenance tracking system for Python data science scripts. Vamsa relies on the aforementioned design principles: (1) it uses a variety of new and existing static analysis techniques that do not make any assumption about the structure of the script and do not require any user input (2) its core modules are agnostic of the ML libraries invoked in the script. To achieve that, Vamsa queries an external knowledge base containing APIs of various ML libraries.  It can thus operate on all kinds of data science scripts as long as the knowledge base has been updated to contain the appropriate APIs. This design allows improving coverage by simply adding more ML APIs in the knowledge base without any further code changes in Vamsa or the user scripts.

\item Using data science scripts from Kaggle~\cite{kaggle_api} and publicly available Python notebooks~\cite{rule2018exploration}, we perform experiments using $26K$ scripts and verify the effectiveness of Vamsa in terms of coverage and performance. We also evaluate the accuracy on a smaller subset of manually labeled data. Our analysis shows that Vamsa's precision and recall range from $90.4\%$ to $99.1\%$ and its latency is in the order of milliseconds for average size scripts.

\item We present a real, end-to-end scenario where Vamsa can help debug an ML model deployed in production to predict job slowdowns in clusters of thousands of nodes.

\end{enumerate}

\section{Problem Statement}
\label{sec:pre}

We now define the problem of {\em automated  ML provenance tracking} that Vamsa targets. In essence, given a data science script, our goal is to identify which columns in a dataset have been used to train a particular ML model by analyzing the script during static analysis time---hence, automatically capturing the relationships between data sources and models at a coarse-grained level. More formally:

\stitle{ML Provenance Tracking}. Given a data science script, our goal set is to find all triples $\langle M, D, C \rangle$: each $M \in \mathcal{M}$ is a constructed machine learning model trained in the script using data sources $D$ (e.g., database tables or views, CSVs, spreadsheets, or external files).\footnote{Note that there are also data science scripts that do not perform any model training but provide other functionality (e.g., visualization, optimization, etc.) In this work, we focus on data science scripts that include statements that train ML models as our goal is to capture the relationships between data sources and generated ML models.} ln particular, the model is trained using features (and optionally labels) derived from a subset of columns of data sources $D$, denoted as $C$. The goal is to identify each trained model $M$ in the script, its data sources $D$, and columns $C$ from $D$ that were used to train model $M$.

\begin{example}
\label{ex-problem}
The input to Vamsa is a data science script such as the one in Figure~\ref{fig-motivation-example}. The script reads from \texttt{heart\_disease.csv} as a data source $D$ and trains an ensemble of decision trees using \texttt{catboost}~\cite{prokhorenkova2018catboost}. In this script, only a single model was trained. Note that not all the columns of the data source have been used to derive the model's features and labels.  To select features, a range of columns $[3,+\infty)$ from $D$ is explicitly extracted, followed by the drop of the columns $\{\texttt{SSN},\texttt{Target}\}$. Similarly, only the \texttt{Target} column was used for labels. Thus, the desired output is a triple $\langle M, D, C \rangle$ with $M=\{\texttt{clf}\}$ is the variable that contains the trained model, $D=\{\texttt{heart\_disease.csv}\}$ is the training dataset, and $C$ is the set $[3,+\infty) - \{\texttt{SSN}\}$. The goal of Vamsa is to analyze the script and produce this output.
\end{example}

Note that under the static analysis setting the problem is undecidable. In this work, we will not focus on sources of undecidability (e.g., conditionals and loops) because their presence in data science scripts is minimal~\cite{dsonds} (Section~\ref{sec:conclusion} discusses future directions to account for such cases).  Finally note that, in this work, we focus on scripts written in Python because this is the major language currently used by data scientists~\cite{KaggleSurvey,KaggleFigure,dsonds}. The principals behind our techniques, however, naturally generalize to other languages as well.

\section{Vamsa Architecture}
\label{sec:arch}

Vamsa is a system designed to address the ML provenance tracking problem as described in the previous section. To do so, we have built Vamsa based on the following two design principles:

\noindent \textbf{1. Support for unmodified Python scripts}: Unlike other works that require the user to add logging information in their script using appropriate APIs~\cite{miao2017modelhub,mlflow}, our goal was to build a system that does not require any user input or changes to the user code as it is a manual and error-prone process. Moreover, there is already a large corpus of existing production scripts and adjusting them to track provenance information might not be feasible. 

\noindent \textbf{2. Ability to incorporate new ML libraries and frameworks}: Data science is an evolving field, with new frameworks coming up and existing frameworks evolving their APIs.
An ML provenance tracking system should be designed to accommodate all users and their needs and should not be restricted to a specific set of ML libraries. Thus, the algorithms used by the system should be agnostic of the ML frameworks invoked in the script.

Building such a system is challenging. On the one hand, user input in any form is not acceptable. This is because: (1) user input is a manual process--hence, expensive and error-prone---and (2) data scientists that author such scripts are not proficient with neither provenance tracking nor the, often limited, provenance tracking APIs.  On the other hand, support of various kinds of ML libraries that data scientists use is needed. To strike a balance between these two relatively conflicting objectives, Vamsa: (1) analyzes Python scripts using existing and novel static analysis techniques that are agnostic of the particular ML libraries invoked (or their different versions) but (2) relies on an external knowledge base of APIs to collect semantic information about the script when needed. 

\begin{figure}[t!]
\centering
\centerline{\includegraphics[width=\columnwidth]{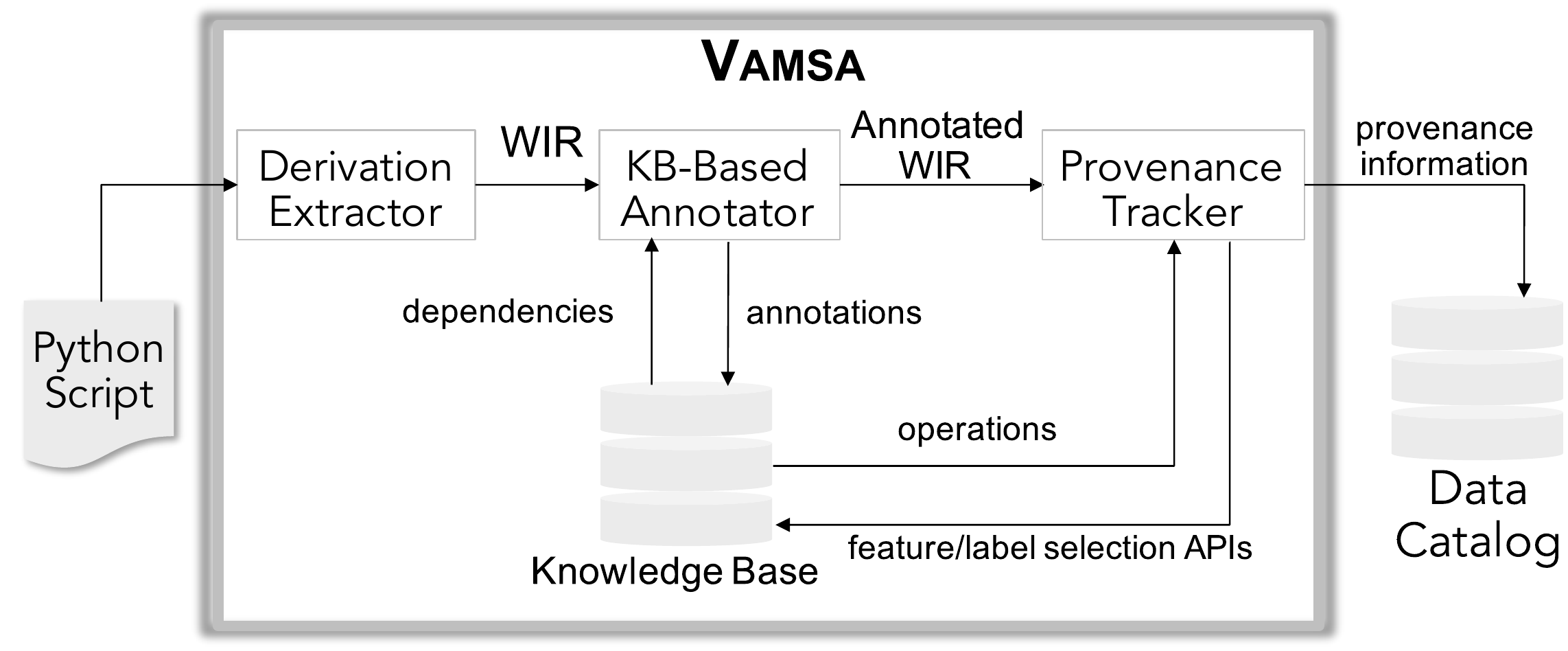}} 
\vspace{-1em}
\caption{{\small Vamsa Architecture.}}
\label{fig-arch}
\vspace{-1em}
\end{figure}

Figure~\ref{fig-arch} illustrates the architecture of Vamsa. At a high-level, Vamsa performs static analysis on the input Python script to determine the relationships between all the variables in the script, followed by an annotation phase that assigns semantic information to the variables
in the script. During this phase, Vamsa queries the external knowledge base to obtain information about the semantics of various ML operations. %
It then uses a generic provenance tracking algorithm that extracts the feature set for all the ML models trained in the script and stores this information in a central catalog that can be accessed by various provenance applications.

More specifically, Vamsa processes data science scripts with the following three major modules: the \textit{Derivation Extractor}, the \textit{KB-based Annotator}, and the \textit{Provenance Tracker}. Next we give a brief overview of these components, and point forward to individual sections for in-depth discussion on each component.

\noindent (1) The \textit{Derivation Extractor} generates a workflow intermediate representation (WIR) of the script by extracting major workflow elements including imported libraries, variables, functions, 
and their dependencies using standard static analysis techniques (Section~\ref{sec:wir}). 

\noindent (2) The \textit{KB-based Annotator} annotates variables in WIR based on their roles in 
the script (e.g., features, labels, and models). To this end, it uses our generic annotation algorithm and a pluggable
knowledge base that contains information about the various APIs of different ML libraries. Through the knowledge base, we are able to declaratively introduce semantic information for operations in the Python script. This design allows us to operate on all kinds of Python libraries (ML or otherwise) as long as the appropriate APIs are included in the knowledge base. It also allows users to improve coverage by simply adding the APIs of more libraries in the knowledge base, without having to modify their code or the components of Vamsa. (Section~\ref{sec:annotate})

\noindent (3) The \textit{Provenance Tracker} infers a set of columns that were explicitly included in or excluded from the features/labels by using the annotated WIR and consulting the knowledge base.  The Provenance Tracker is able to operate in both supervised and unsupervised learning settings. In the former case it tracks both features and labels, while in the latter it tracks only features. (Section \ref{sec:tracker})

\eat{We remark that acquiring labeled data is non-trivial or even infeasible~\cite{griffon} in real-world settings.}

\eat{\vspace{1ex}
Vamsa does not make any assumption about the ML libraries/frameworks used to train the models. By utilizing a modular architecture combined with a knowledge base of APIs for various ML libraries, Vamsa is able to operate on all kinds of Python libraries as long as the appropriate APIs are included in the knowledge base. Additionally, this design allows users to improve coverage by simply adding more ML APIs in the knowledge base, without having to modify their code or Vamsa's other components.

To evaluation Vamsa, %
we %
have populated our knowledge base with  APIs from four well-established data science libraries: scikit-learn~\cite{pedregosa2011scikit}, XGBoost~\cite{xgboost}, LightGBM~\cite{lightgbm}, and Pandas~\cite{mckinney2011pandas}. Nevertheless, Vamsa can operate on top of any other library such as CatBoost~\cite{prokhorenkova2018catboost}, StatsModels~\cite{seabold2010statsmodels}, and Graphlab~\cite{graphlab}, among others. 
}

\section{Derivation Extractor}
\label{sec:wir}

As discussed previously, the derivation extractor uses standard static analysis techniques to parse the Python script, build a {\em workflow model} which captures the dependencies among the elements of the script including imported libraries, input arguments, operations that change the state of the program, and the derived output variables. This model is captured in a \textit{workflow intermediate representation} (WIR). In this section, we will not analyze in depth these techniques, as they rely on well-known concepts~\cite{engineeringcompiler}, but will provide important notation and background required for subsequent components. In~\cite{tr} we present in detail these techniques.

\stitle{Operations}. We denote the set of all variables in the data science script as $V$. An operation $p \in P$ operates on an ordered set of input variables $I$ to change the state of the program and/or to derive an ordered set of output variables $O$. An operation may be called by a variable, denoted as caller $c$. While an operation may have multiple inputs and outputs, it has at most one caller.

\begin{example}
\label{ex-proc}
In Figure~\ref{fig-motivation-example}, the import statements, \texttt{read\_csv($\cdot$)} in line~4, attribute \texttt{values} in line~5, \texttt{CatBoostClassifier($\cdot$)} in line~9, and \texttt{fit($\cdot$)} in line~10 are examples of operations. Consider the \texttt{fit($\cdot$)} operation: it is invoked by the \texttt{clf} variable and takes three arguments namely, \tfm{} and \tgt, and an evaluation set. While \texttt{fit($\cdot$)} does not explicitly produce an output variable, it changes the state of the variable \texttt{clf}.
\end{example}

\begin{figure}[tb!]
\centering
\centerline{\includegraphics[width=\columnwidth]{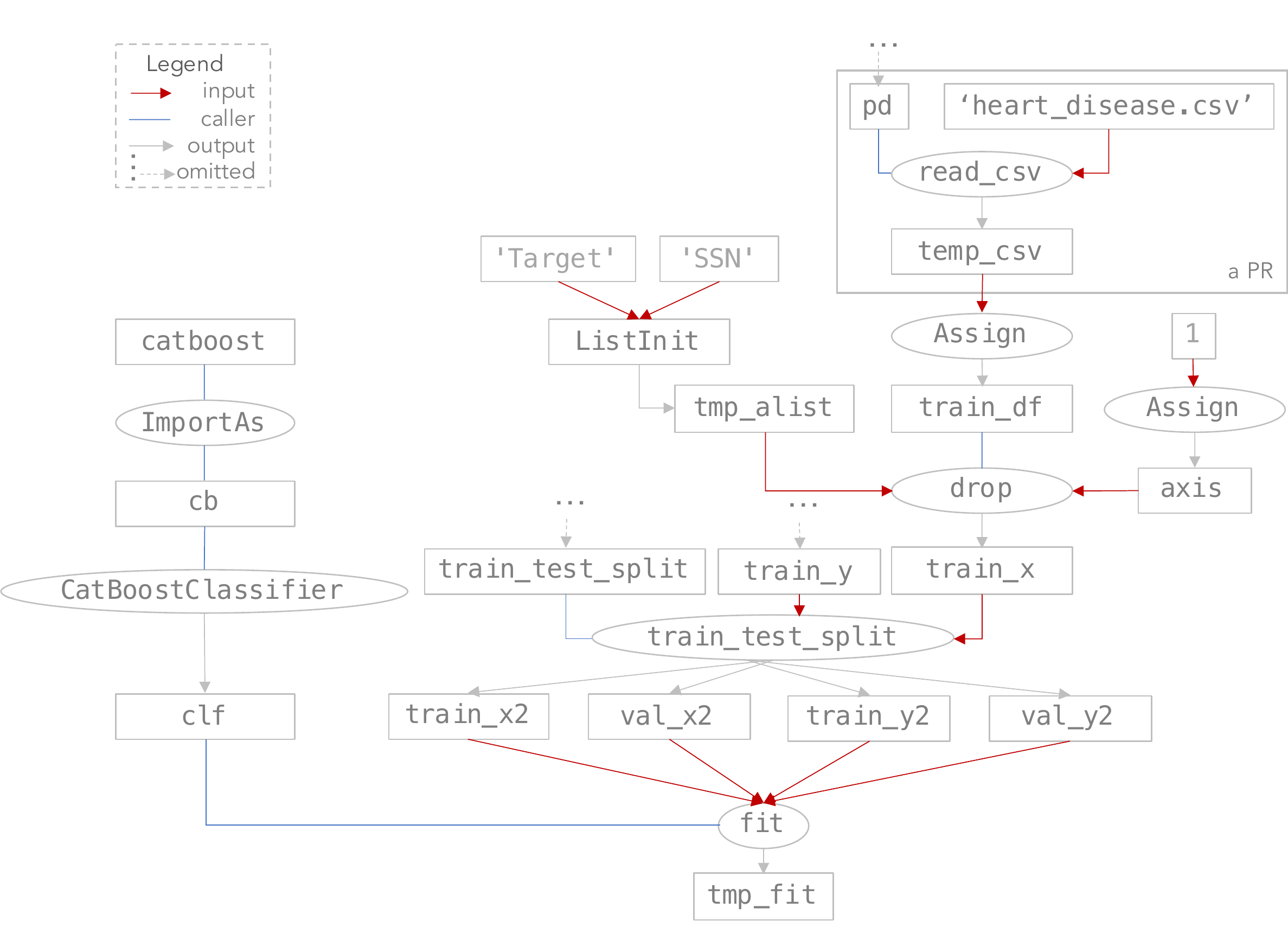}}
\vspace{-1ex}
 \caption{{\small An example WIR.}}
\label{fig-wir}
\vspace{-2em}
\end{figure}

\stitle{Provenance relationship.} An invocation of an operation $p$ (by an optional caller $c$) %
depicts a {\em provenance relationship} (\LR). A \LR is represented as a quadruple $(I, c, p, O)$, where $I$ is an ordered set of  
input variables, (optional) variable $c$ refers to the caller object, $p$ is the operation, and $O$ is an ordered set of output variables that was derived from this process. %
A \LR can be %
represented as a labeled directed graph, 
which includes (1) a set of \textit{input edges} (labeled as \textit{`input\_edge'}), 
where there is an input edge $(v,p)$ for each $v \in I$, %
(2) a \textit{caller edge} (labeled as \textit{`caller\_edge'}) %
$(c,p)$ if $p$ is called by $c$, and 
(3) a set of \textit{output edges} (labeled as \textit{`output\_edge'}), 
where there is an output edge $(p,v)$ for each $v \in O$.  
For consistency, we create a temporary output variable for the operations that do not explicitly generate one.

\begin{example}
\label{ex-lineage}
Consider line~4 in Figure~\ref{fig-motivation-example} where the CSV file `heart\_disease.csv' is read. The corresponding \LR is depicted in Figure~\ref{fig-wir} (rectangle in top right) and corresponds to the quadruple $(I, c, p, O)$ where $I=\{\langle \texttt{heart\_disease.csv} \rangle\}$, $c=\texttt{pd}$, and $p=\texttt{read\_csv}$. Finally, we create a temporary variable \texttt{tmp\_csv} and set $O=\{\texttt{tmp\_csv}\}$ to used as the input to other \LRs. 
\end{example}

\stitle{Workflow Intermediate Representation.} \LRs are composed together to form a WIR $G$, 
which is a directed graph that represents the sequence and dependencies among the extracted \LRs. The WIR is useful to answer queries such as: ``Which variables were derived from other variables?'', `` What type of libraries and modules were used?,''  and ``What operations were applied to each variable?''. More formally, a WIR is a directed bipartite graph $G=(V \cup P, E)$ with vertices $V \cup P$ and edges $E \subseteq (V \times P) \cup (P \times V)$. Each edge has a type drawn from the set: \{input\_edge, output\_edge, caller\_edge\}.

\begin{example}
\label{ex-wir}
Figure~\ref{fig-wir} illustrates a fraction of the WIR generated for the script of Figure~\ref{fig-motivation-example}. The variables and operations are represented by rectangles and ovals, respectively. The caller; input; and output edges are marked in blue; red; and black color, respectively. Consider the operation \texttt{fit}. One can tell the following from the WIR: 1) it is called by variable \texttt{clf}; 2) it has two ordered input variables \texttt{train\_x2} and \texttt{train\_y2}; and 3) a temporary variable, denoted as \texttt{tmp\_fit}, was created by Vamsa as its output.
\end{example}

Vamsa generates workflows using standard data flow analysis techniques~\cite{engineeringcompiler}. More specifically, it first parses the script to obtain a corresponding \textit{abstract syntax tree (AST)}~\cite{ast_page, ast_more_info} representation. Then, it identifies the relationships between nodes of the AST to generate  \LRs. Finally, it composes the generated \LRs into the WIR~\cite{tr}.

\begin{table}[t]
\scriptsize
\setlength\tabcolsep{1pt}
\begin{center}
\begin{tabular}{|c|c|c|c|c|c|}
\hline
\textbf{Library} & \textbf{Module}  & \textbf{Caller} & \textbf{API\_Name} & \textbf{Inputs}                                                                  & \textbf{Outputs}                                                                                \\ \hline
catboost         &    NULL              &  NULL                & CatBoostClassifier                                                                                         & eval\_metrics: \hp & model \\ \hline
catboost         &  NULL                 & model           & fit                & \begin{tabular}[c]{@{}c@{}c@{}}\tfm \\ \tgt \\eval\_set: validation sets \end{tabular} & trained model                                                                                   \\ \hline
sklearn          & model\_selection & NULL                & train\_test\_split & \begin{tabular}[c]{@{}c@{}c@{}}\tfm\\ \tgt \\test\_size: testing ratio  \end{tabular} & \begin{tabular}[c]{@{}c@{}}\tfm\\ validation features\\ …\end{tabular} \\ \hline
\end{tabular}
\end{center}
\caption{Example of facts in Vamsa knowledge base}
\label{tbl:kb_tuples}
\vspace{-2em}
\end{table}

\section{KB-based Annotator}
\label{sec:annotate}

The generated WIRs capture the dependencies among the variables and operations in a script, as we discussed in the previous section. Unfortunately, WIRs alone do not provide {\em semantic information} such as the role of variables in a script (e.g., ML model or features) or the type of objects (e.g., CSV file or dataframe). To address the ML provenance tracking problem, however, this semantic information needs to be included in the output provenance information. To this end, the goal of the KB-Based Annotator, that we focus on in this section, is to annotate variables in WIRs with semantic information.

Finding the role of each variable in a WIR is a challenging task for multiple reasons. First, one cannot accurately deduce the type of inputs and outputs of an operation by only looking at the name of the operation. This is because different ML libraries may use the same operation name for different tasks. Second, even in the same library, an operation may accept different number of inputs or provide different outputs. For example, the \texttt{fit} function of \texttt{sklearn} ~\cite{pedregosa2011scikit} can accept one or two inputs depending on the task (e.g., one for clustering or two for classification or regression). Finally, some variables (e.g., feature sets) are hard to semantically annotate early on. For instance, we cannot decide whether the returned dataframe from a \texttt{read\_csv} call of \texttt{pandas} is a training set, before in itself (or after preprocessing steps) becomes input to a training function. 

Besides the technical challenges outlined above, a semantic annotation framework to be usable across various data science scripts must be: (1) compatible with the various ML libraries and their different versions, and (2) extensible to accommodate new libraries. 

To this end, we propose a generic annotation algorithm (Section~\ref{sec:annotation_algo}) that is \textit{agnostic} of the underlying ML libraries used in the script by querying an external knowledge base of ML APIs (\kb) when semantic information is needed (Section~\ref{sec:kb}) and is able to propagate annotations across different elements of the WIRs. \eat{The \kb can be used to answer questions such as ``What is the role of the input variables of a particular operation belonging to a given ML library?''. }

Note that there is already a substantial effort from the Python community to annotate various libraries and their external APIs with type information for static analysis purposes~\cite{typeshed}. As these initiatives mature, the population of the \kb that Vamsa queries will become straightforward.  In our current prototype, and similar to other efforts for \kb population~\cite{ives2019dataset}, the construction of the \kb is manual. As we show in our experiments, however, our manual (yet minimal) \kb results in large coverage on big collections of data science scripts. This is primarily because many data science scripts rely on similar coding patterns~\cite{dsonds}.

\begin{figure}[tb!]
\begin{center}
{\small
\begin{minipage}{3.36in}
\myhrule 
\vspace{-1ex}
\mat{0ex}{
{\bf Algorithm}~\annotAlg \\
\sstab {\sl Input:\/} \= WIR $G$ and knowledge base \kb.\\ 
{\sl Output:\/} Annotated WIR $G^+$. \\
\bcc \hspace{1ex}\= Find the \kw{Import} process nodes in $G$ as the seed set $\S$; \\
\icc\> \ForEach $v_s \in \S$ \Do \\
\icc\> \hspace{1ex} Extract library $L$ and module $L'$; \\
\icc\> \hspace{1ex} Starting from $v_s$, follow a DFS forward traversal on \LRs: \\
\icc\> \hspace{3ex} \ForEach seen $\LR=\langle I, c, p, O \rangle$ \Do \\
\icc\> \hspace{5ex} Obtain annotation of $v_i \in I$ and $v_o \in O$ \\ \> \hspace{5ex} by invoking $\kb(L,L',c, p)$ \\
\icc\> \hspace{5ex} \ForEach annotated $v_i \in I$ \Do  \\ 
\icc\> \hspace{7ex} Starting from $v_i$, follow a DFS backward traversal\\
\> \hspace{7ex} on \LRs:  \\ 
\icc\> \hspace{9ex} \ForEach seen $\LR=\langle I, c, p, O \rangle$ \Do \\
\icc\> \hspace{11ex} Obtain annotation of $v_i \in I$ by invoking $\kb(O, p)$ \\ 

\icc\> \hspace{1ex} \Return $G^+$; 
}
\vspace{-1ex}
\myhrule
\end{minipage}
}
\end{center}
\caption{Annotation algorithm}
\label{fig:annotAlg}
\end{figure}

\subsection{Knowledge Base of ML APIs}
\label{sec:kb}
\eat{The \kb is a standalone database that contains information about ML libraries. For each API in an ML framework, \kb knows its library name and version, its module name (if any), the specific type of its caller object, as well as the data type and user-defined annotation for each of its input/output set.}

The \kb contains information about ML libraries and APIs including, but not limited to, library and operation names; versions; modules; as wells as types of inputs and outputs of operations.

\begin{example}
Table~\ref{tbl:kb_tuples} shows three tuples in our \kb that are utilized by the annotation algorithm to identify the variables that correspond to models and features in the script of Figure~\ref{fig-motivation-example}. The second tuple shows that when the operation \kw{fit} is called via a model constructed by \kw{catboost} library, its first and second input are \tfm{} and \tgt{}, respectively. It also accepts the validation sets as input. The output of the operation is a trained model.
\end{example}

To facilitate the annotation of WIR variables, \kb supports two types of queries. The first one denoted as $\kb(L,L',c, p)$  takes as input the name of a library $L$, module $L'$, caller type $c$, and operation $p$ and returns a set of user-defined annotations that describe the role and type for each input/output of operation $p$. The second one denoted as $\kb(O, p)$ obtains the annotations of the input variables of operation $p$ given the annotations of its output $O$.

\subsection{Annotation Algorithm}
\label{sec:annotation_algo}

The annotation algorithm traverses the WIR and annotates its variables by querying the \kb when needed. After each annotation, new semantic information about a WIR node is obtained that can be used to enrich the information associated with other WIR variables, as is typical in the analysis of data flow problems~\cite{engineeringcompiler}. The propagation of semantic information is achieved through a combination of forward and backward traversals of the WIR. 

The algorithm (Figure~\ref{fig:annotAlg}) starts by finding a set of \LRs with $p=\texttt{Import}$ as a seed set $\S$ for upcoming DFS traversals (line~1). These \LRs contain the information about imported libraries and modules in the Python script. For each $v_s \in \S$, the algorithm extracts the library name $L$ and the potentially utilized module $L'$ (line~3). It then initiates a DFS traversal (line~4) that, starting from $v_s$, traverses the WIR in a forward manner (i.e., by going through the outgoing edges). For each seen \LR, it obtains the annotation information for both of its inputs $I$ and outputs $O$ by querying the knowledge base (lines~5-6) as described in the previous section. 

If a new annotation was found for an input variable $v_i \in I$, the algorithm initiates a backward  DFS traversal. 
As the input variable $v_i$ can be the output of another \LR, for new information discovered for $v_i$, we can propagate this information to other \LRs in which $v_i$ is their output. In particular, starting from $v_i$, the algorithm traverses the WIR in a backward manner (i.e., by going through the incoming edges) (line~8). During the backward traversal, the \kb is used to obtain information about the inputs of an operation given its already annotated output. In each initiated DFS traversal, each edge is visited only once. The algorithm terminates when we cannot obtain more information from initiating more forward or backward traversals~\cite{tr}.

\begin{example}
\label{ex-annotation}
Operating on the WIR of Figure~\ref{fig-wir}, the annotation algorithm initializes the seed set $\S$ with one \texttt{import} operation and sets $L=$ 
 \texttt{catboost} and $L'=\perp$. Once it visits the
 $p$ = \texttt{CatBoostClassifier} operation, it queries the \kb to obtain the annotation of its output. Given, $L$, $L'$, $c=$ \texttt{catboost} and $p$, the \kb annotates \texttt{clf} as a model. Since there exists no input edge for the \texttt{CatBoostClassifier} node in this WIR, no backward traversal is initiated. The algorithm moves forward and visits the \texttt{fit} function. It queries the \kb with the same $L$ and $L'$, but updated $c=model$ and $p=\texttt{fit}$. The algorithm annotates the output of $\texttt{fit}$ as trained model and then stops the forward propagation since there are no more outgoing edges in the node. However, at this time, \kb successfully annotated the \texttt{train\_x2} and \texttt{train\_y2} as the \tfm{} and \tgt{}, respectively. Thus, two backward traversals are started to propagate this information as much as possible to the previous nodes in the WIR. Let us follow the DFS that was started from  \texttt{train\_x2}. By visiting the \texttt{train\_test\_split} node, the algorithm annotates \texttt{train\_x} as \tfm{}. Similarly, it back-propagates the new annotation to \texttt{train\_df} as the caller of \texttt{drop} operation. The algorithm continues until we cannot obtain more annotation information. 
\end{example}

\section{Provenance Tracker}
\label{sec:tracker}

\begin{figure}[tb!]
\centering
\centerline{\includegraphics[width=.8\columnwidth]{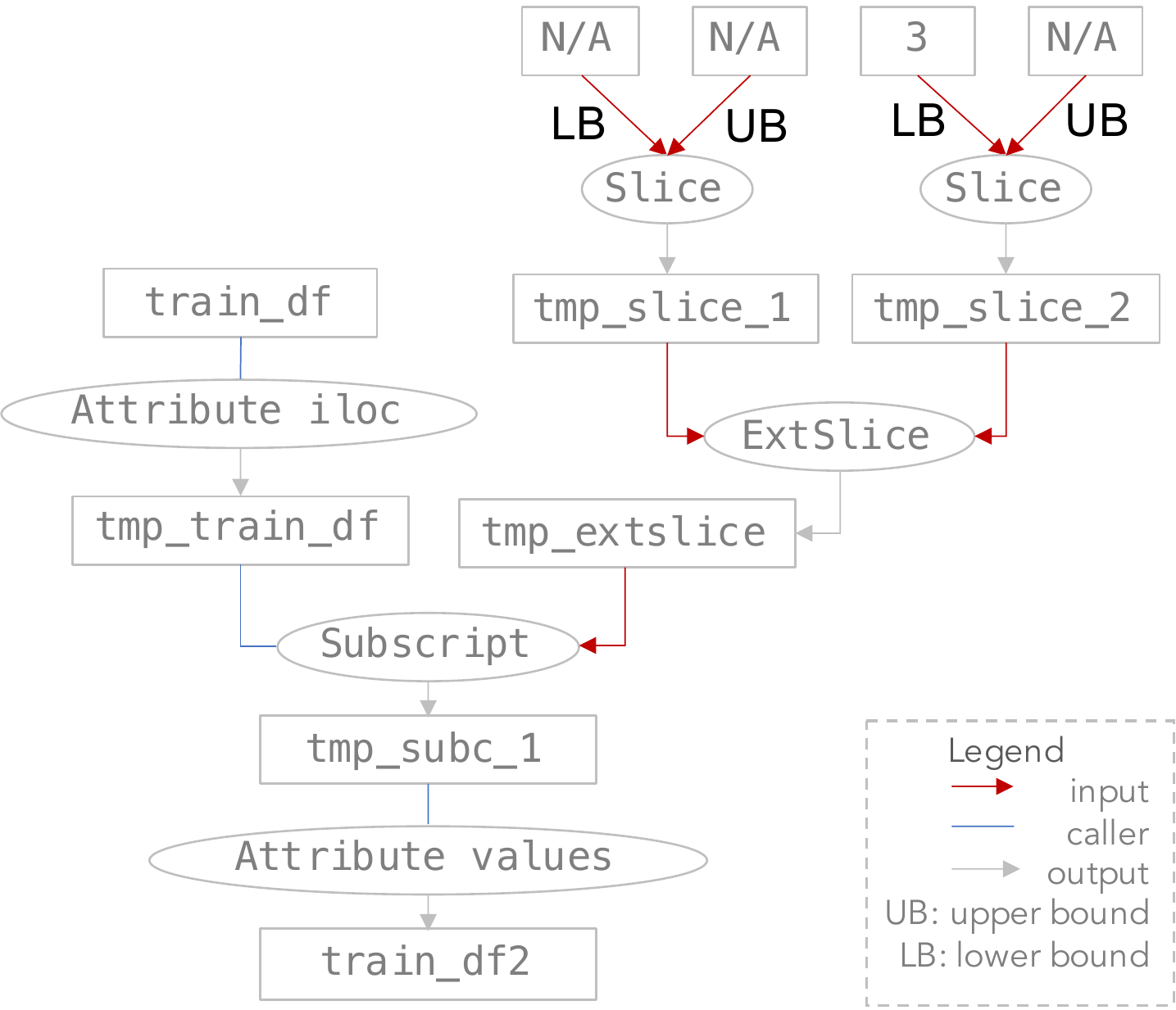}}
 \caption{{\small WIR with \texttt{Subscript} operation}}
\label{fig-wir-subscript}
\vspace{-1em}
\end{figure}

We next introduce the Provenance Tracker component of Vamsa. The provenance tracker takes as input the WIR annotated (i.e., the output of the algorithm in Figure~\ref{fig:annotAlg}) and its goal is to automatically detect the subset of columns in a data source that was used to train a ML model and, as such, forms the overall output of Vamsa.

To identify the columns, we need to investigate the operations in the annotated WIR that are connected to variables that contain \tfm{} and \tgt{} in their annotation set. Note that there are various operations that take \tfm{} (or \tgt{}) as their caller or input, and may apply transformations on them (e.g., drop a set of columns, select a subset of rows upon satisfaction of a condition, or copy into other variables) that the Provenance Tracker needs to account for.

\eat{
To this end, we enrich our \kb with a new table to guide the Provenance tracker. The new table consists of two types of operations: operations that may exclude or include columns. Furthermore, these operations can be classified as 1) operations from various Python libraries that exclude columns (e.g., \texttt{drop} and \texttt{delete} in \texttt{Pandas}) or explicitly select a subset of columns (e.g., \texttt{iloc} and \texttt{ix}), and 2) a few native Python operations (e.g., \texttt{Subscript}, \texttt{ExtSlice}, \texttt{Slice}, \texttt{Index}, and \texttt{Delete}~\cite{ast_page, ast_more_info}). Each entry in this table is annotated with a flag \texttt{column\_exclusion}$=$\texttt{True}, if the corresponding operation can be used for column exclusion and a \kw{traversal\_rule} (i.e., a function specifying how to start a backward traversal from the node's input edges in order to identify a set/range of indices/column names). 
}

To this end, we enrich our \kb with a new table to guide the Provenance Tracker. The new table follows the structure of Table~\ref{tbl:kb_tuples}. This time, however, each entry is annotated further with two attributes: \texttt{column\_exclusion} and \kw{traversal\_rule}. The \texttt{column\_exclusion} attribute is set to True, if the operation explicitly excludes columns (e.g.,  \texttt{drop} and \texttt{delete} in \texttt{Pandas}), and False otherwise. The \kw{traversal\_rule} is a function specifying how to start a backward traversal from the node's input edges in order to identify a set/range of indices/column names. Example~\ref{ex-wir-subscript} provides an example to clarify these notions. Finally, note that we query this table by invoking $\kb_{C}(p)$ where $p$ is the operation. The query returns $\emptyset$, if there is no matching entry in the \kb, or the \kw{column\_exclusion} and \kw{traversal\_rule}, otherwise.

\eat{
consists of two types of operations: operations that exclude or not columns. Operations that exclude columns can be classified as 1) operations that directly exclude columns (e.g., \texttt{drop} and \texttt{delete} in \texttt{Pandas}) or 2) explicitly select a subset of columns (e.g., \texttt{iloc} and \texttt{ix} in Pandas). Each entry in this table is annotated with a flag \texttt{column\_exclusion}$=$\texttt{True}, if the corresponding operation can be used for column exclusion and a \kw{traversal\_rule} (i.e., a function specifying how to start a backward traversal from the node's input edges in order to identify a set/range of indices/column names). 
}

\begin{example}
\label{ex-wir-subscript}
Figure~\ref{fig-wir-subscript} is another fraction of the WIR that was generated from line~5 of the script in Figure~\ref{fig-motivation-example} that includes a \texttt{Subscript} operation. The statement in line 5 keeps all the rows but only includes the columns from index $3$ to the last index in the dataset. One can find the set of included columns by traversing backward the nodes following the input edge of the \texttt{Subscript} operation and reaching the constant values connected with the \texttt{Slice} operations. The traversal rule associated with the \texttt{Subscript} operation in our KB indicates that the input edge of this node must be followed in a backward manner to eventually reach the selected columns. Note that this is the case for all WIRs that contain this operation.
\end{example}

The overall provenance tracking algorithm is illustrated in Figure~\ref{fig:fTracker}. The algorithm takes as input the annotated WIR $G^+$ and the \kb, and returns two column sets: (1) columns from which features were explicitly derived  (denoted as \textit{inclusion set} $C^+$) and (2) columns that are explicitly excluded from features (denoted as \textit{exclusion set} $C^-$). The algorithm scans each \LR to find the ones with a variable that has been annotated as \tfm{} and an operation which can be used for feature selection based on the information stored in the \kb (line~2-3). 
Note that the exact same logic is used to derive inclusion and exclusion sets for labels, as opposed to features, and we combine the two in a single pass over the PRs of the WIR $G^+$.

A core component of the algorithm is the ~\guideEval operator (shown in Figure~\ref{fig:fTracker}) that starts a guided traversal of the WIR based on the information in the \kb. For each of the selected \LR, the \guideEval operator queries the \kb and obtains the corresponding \kw{column\_exclusion} flag and \kw{traversal\_rule} (line~1).  The operator checks if this \LR contains constant values in its input set (line~2). If so, it incorporates the discovered constant values/range of column indices into the inclusion/exclusion sets. In case the \LR does not directly contain the columns, the \guideEval operator follows the \kw{traversal\_rule} to obtain a new \LR on $G^+$ (line~7) that needs to be evaluated. It then calls the \guideEval operator again for this \LR (line~8).
\begin{figure}[tb!]
\begin{center}
{\small
\begin{minipage}{3.36in}
\myhrule 
\vspace{-1ex}
\mat{0ex}{
{\bf Algorithm}~\texttt{PTracker} \\
\sstab {\sl Input:\/} \= Annotated WIR $G^+$, knowledge base \kb.\\ 
{\sl Output:\/} Column inclusion set $C^+$, column exclusion set $C^-$. \\
\bcc \hspace{2ex}\= $C^+$ $:=$ $\emptyset$; $C^-$ $:=$ $\emptyset$; \\
\icc\> \ForEach \LR \IN \LRs \Do \\
\icc\> \hspace{2ex} \If it has a variable that was annotated as \tfm{} or \tgt{}\\
\hspace{6ex} \And $\kb_{C}(p)\neq\emptyset$ \\
\icc\> \hspace{6ex} $\guideEval(\LR, G^+,\kb, C^+, C^-)$ ;\\ 
\icc\> \Return $C^+, C^-$; 
}
\mat{0ex}{
{\bf Operator}~$\guideEval(\LR, G^+,\kb, C^+, C^-)$ \\
\sstab {\sl Input:\/} \=  Visited \LR, annotated WIR $G^+$, knowledge base \kb,\\
 column inclusion set $C^+$, column exclusion set $C^-$. \\
{\sl Output:\/} Updated $C^+$, $C^-$. \\
\bcc\> \kw{condition}, \kw{column\_exclusion}, \kw{traversal\_rule} = $\kb_{C}(p)$;\\
\icc\> \If \LR has constant inputs $\kw{cnst}$ \Then \\
\icc\> \hspace{2ex}  \If \kw{column\_exclusion}=\kw{True} \Then\\
\icc\> \hspace{4ex}  $C^- := C^- \cup \kw{cnst}$;\\ 
\icc\> \hspace{2ex}  \Else $C^+ := C^+ \cup \kw{cnst}$; \\
\icc\> \hspace{2ex}  \Return $C^+, C^-$; \\
\icc\> Obtain new \LR on $G^+$ based on \kw{traversal\_rule};\\
\icc\> $\guideEval(\LR, G^+,\kb, C^+, C^-)$; \\
}
\vspace{-5ex}
\myhrule
\end{minipage}
}
\end{center}
\vspace{-3ex}
\caption{Provenance tracking algorithm} \label{fig:fTracker}
\vspace{-2em}
\end{figure}

\begin{example}
\label{ex-wir-subscript2}
Continuing with example~\ref{ex-wir-subscript}, the provenance tracking algorithm finds the \texttt{drop} operation with a caller that was annotated as \tfm{} (Figure~\ref{fig-wir}) and thus invokes the \guideEval operator.Based on the information in the \kb, we know that the operation was used for feature exclusion. Thus, the algorithm follows the traversal rule to perform a backward traversal from its the operation's input edge until it finds the constants \texttt{`Target'}, and \texttt{`SSN'}. These two columns are then added to the exclusion set. When the feature tracking algorithm finds the \texttt{Subscript} operation (Figure~\ref{fig-wir-subscript}) in the annotated WIR, it invokes the \guideEval operator again. The operator only obtains the corresponding traversal rule from the \kb and initiates a backward traversal starting from the input edge of the \texttt{Subscript} operation. A similar process is performed when the \guideEval operator visits \texttt{ExtSlice} and \texttt{Slice} nodes. Using the traversal rule for \texttt{Slice}, for instance, the algorithm looks for a range of columns with lower bound (respectively upper bound) that can be found by traversing the appropriate input edges of the \texttt{Slice} node (see Figure~\ref{fig-wir-subscript}).
\end{example}

\section{Experimental Evaluation}
\label{sec:exp}

\eat{\begin{table}[]
\begin{center}

\begin{tabular}{|c|c|c|}
\hline
\textbf{Dataset}  & \textbf{\begin{tabular}[c]{@{}c@{}}Error-free\\ \& Python 3 \\ compatible\\ $\ntbk_2$/$\kaggle_2$\end{tabular}}  & \textbf{\begin{tabular}[c]{@{}c@{}}Scripts with \\selected \\ ML libraries\\ $\ntbk_3$/$\kaggle_3$\end{tabular}} \\ \hline
\ntbk           ($807K$)       & $447K$                                                                                                                                                                   & $28.9K$                                                                              \\ \hline
\kaggle         ($4.8K$)       & $4.2K$                                                                                                                                                                     & $1.2K$                                                                               \\ \hline
\end{tabular}
\caption{Output of pre-processing pipeline}
\label{tbl-datasets}
\end{center}
\end{table}}

We now evaluate Vamsa on large corpora of Python scripts and provide an analysis of our experimental results. We also present an end-to-end scenario that we encountered in production to better show how Vamsa can facilitate model debugging.

\begin{table*}[tb]

\begin{center}
\begin{tabular}{|c|c|c|c|c|c|c|c|c|}
\hline
\multirow{2}{*}{\textbf{Dataset}} & \multicolumn{2}{c|}{\textbf{Feature Exclusion}}                & \multicolumn{2}{c|}{\textbf{Feature Inclusion}}    & \multicolumn{2}{c|}{\textbf{Label Inclusion}}            & \multicolumn{2}{c|}{\textbf{Annotation Precision}} \\ \cline{2-9} 
                         & \textbf{Precision} & \textbf{Recall} &  \textbf{Precision} & \textbf{Recall} &   \textbf{Precision} & \textbf{Recall}  &\textbf{Model}        & \textbf{Train Dataset}        \\ \hline
\textbf{$\kaggle$}     & $99.11\%$            & $97.72\%$                          & $91.37\%$            & $92.89\%$        &$95.47\%$  &  $95.67\%$              & $100\%$                 & $99.33\%$                    \\ \hline
\textbf{$\ntbk$}       & $94.83\%$            & $96.58\%$                        & $90.54\%$            & $94.08\%$                &$90.36\%$  &  $90.78\%$         & $100\%$                & $98.66\%$                    \\ \hline
\end{tabular}
\caption{Accuracy of Vamsa on the labeled datasets}
\label{tbl-end-to-end}
\end{center}
\vspace{-2em}
\end{table*}
\subsection{Experimental Settings}

\stitle{Datasets}. To evaluate Vamsa on a variety of data science scripts, we downloaded a large set of  Python scripts from two data sources: (1) a corpus of Python notebooks published in 2017  that was crawled from \kw{Github}~\cite{rule2018exploration} (\ntbk dataset) and (2) a set of Python scripts that we downloaded via the public Kaggle API~\cite{kaggle_api}(\kaggle dataset).
We filtered these corpora to account only for scripts that include \texttt{import} statements, do not have syntax errors, and are compatible with Python 3 (the Python version that Vamsa's implementation currently targets). After applying these filters, we kept the scripts that included strings (e.g., fit) that popular ML frameworks (e.g., scikit-learn, XGBoost~\cite{xgboost}, and LightGBM) in our KB use to train ML models. We selected these ML frameworks because they are among the most common for training ML models~\cite{dsonds}. Note that it is easy to extend to other libraries by just populating the \kb (no code changes are required). The resulting datasets are denoted as \ntbk ($24.8K$ scripts) and \kaggle ($1.2K$) scripts.

\stitle{Experimental methodology.} A challenge when evaluating Vamsa with such large-scale corpora is to determine the correctness of the output. Unfortunately, due to the novel nature of ML provenance tracking, there is no public benchmark available. The brute force approach would be to manually go over the corpus and determine the relationships between ML models and data sources so that we can evaluate Vamsa's output. Since this is not feasible at the scale we are operating, we decided to perform two classes of experiments. First, we select a small subset of scripts for which we manually extract the provenance information (ground truth) and evaluate the accuracy of Vamsa on those. The second class of experiments is performed on the large corpus. The goal is to evaluate the coverage of the system, defined as how often Vamsa extracts the provenance information. We also evaluate Vamsa's efficiency in terms of latency. 

\stitle{Hardware and software configuration.} We conducted our experiments on a Linux machine powered by an Intel $2.30$ GHz CPU with $8$ GB of memory. For all the experiments we used Python $3.7.2$. 

\eat{\begin{table}[]
\begin{center}
\fontsize{8}{8}\selectfont
\begin{tabular}{|c|c|c|c|c|}
\hline
\multirow{2}{*}{\textbf{Dataset}} & \multirow{2}{*}{\textbf{\begin{tabular}[c]{@{}c@{}}Derivation Extractor\\ Coverage\end{tabular}}}&  \multicolumn{2}{c|}{\textbf{KB-based Annotator Coverage}} & \multirow{2}{*}{\textbf{\begin{tabular}[c]{@{}c@{}}Provenance Tracker\\ Coverage\end{tabular}}} \\ \cline{3-4}
                                                   &                 & \textbf{Model}       & \textbf{Train Dataset}       &                                                                                              \\ \hline
\textbf{$\kaggle$}                     &     $89.69\%$                & $94.62\%$                 & $82.91\%$                       & $80.61\%$                                                                                         \\ \hline
\textbf{$\ntbk$}                       &       $97.08\%$                &  $88.52\%$               & $77.81\%$ & $70.20\%$                                                                                         \\ \hline

\end{tabular}
\caption{Vamsa coverage in large-scale evaluation}
\label{tbl-success-fail}
\end{center}

\end{table}}

\subsection{Experiments with Labeled Datasets}
\label{sec:accuracy}

These experiments evaluate the accuracy of Vamsa on a set of Python scripts for which we have manually extracted the relationship between data sources and ML models. From each of the $\kaggle$ and $\ntbk$ datasets, we randomly selected 150 scripts, ensuring that Vamsa can produce
output for all the selected scripts.  We evaluate the accuracy of Vamsa for both features and labels under column exclusion and inclusion using two metrics: precision and recall. The precision shows the proportion of discovered included/excluded columns that were truly included/excluded columns. The recall shows the proportion of the true included/excluded columns that were discovered by Vamsa  to the actual included/excluded columns.

We further investigate how often Vamsa correctly identifies which variables correspond to ML models and which to training datasets as this is a prerequisite for correctly identifying features and labels. To this end, we also report results that show the precision of the
annotation phase (for both models and training datasets).

Table~\ref{tbl-end-to-end} shows the results on the two datasets. For each metric, we report the average values obtained over the $150$ scripts of the dataset. As shown in the table, Vamsa achieves high precision and recall values for all the tasks evaluated. Overall, we can make the following observations:
\begin{enumerate}[leftmargin=*,partopsep=1ex,parsep=1ex]
\item When Vamsa identifies a model, its training dataset, and the corresponding features, the output is highly reliable.
\item Vamsa reported models $100\%$ accurately and made a few mistakes in detecting their training datasets. We further investigated these scripts and found that the data scientists appended the testing data to the training data in order to perform global value transformations. The merged test data then got separated via a slicing operation immediately before training. Vamsa's annotation algorithm was not able to follow this operation, i.e merge followed by split, and mistakenly identified the testing dataset as the training dataset.

\item Vamsa detects column exclusion sets slightly better than column inclusion ones. This is because, for column exclusion, data scientists typically use a set of specific APIs such as \texttt{drop} and \texttt{pop}, \texttt{del} which can be tracked more easily. Note that we did not evaluate column exclusion for the labels as none of the scripts used these APIs for label selection.
\end{enumerate}

\begin{table}[h!]
\begin{center}
\begin{tabular}{|l|c|c|}
\hline

    &\textbf{$\kaggle$} & \textbf{$\ntbk$} \\ \hline
\textbf{Derivation Extractor}  & $89.69\%$  & $97.08\%$ \\ \hline
\textbf{KB-based Annotator (Model)}  & $91.88\%$  & $96.93\%$ \\ \hline
\textbf{KB-based Annotator (Train Dataset)}  & $85.15\%$  & $88.12\%$ \\ \hline
\textbf{Provenance Tracker}  & $80.85\%$  & $74.48\%$ \\ \hline
\end{tabular}
\caption{Vamsa coverage in large-scale evaluation}
\label{tbl-success-fail}
\end{center}
\vspace{-3em}
\end{table}

\subsection{Large-scale Experiments}
\label{sec:large-scale-exp}

In these experiments, we use a large corpus of Python scripts (full $\ntbk$ and $\kaggle$ datasets). The goal is to evaluate the coverage of various components of Vamsa as well as the efficiency of the system. We also present a detailed analysis of the cases where Vamsa was not able to produce an answer.

\stitle{Derivation Extractor}. First, we evaluate the coverage of Vamsa on generating the workflow intermediate representation. Table~\ref{tbl-success-fail} shows the results. The few cases where Vamsa is not able to produce a WIR are mainly due to Vamsa's current implementation.  In particular, we have not yet covered certain constructs in the Python grammar such as \texttt{DictComp}, \texttt{SetComp}, and \texttt{JoinedStr}. However, we note that incorporating these constructs is solely a matter of extending the implementation and does not require any change in Vamsa's design. 

\stitle{KB-based Annotator}. We investigate how often the annotation algorithm identifies ML models and training datasets. Table~\ref{tbl-success-fail} shows the percentage of the cases where the Annotator can annotate at least one variable as a model and one other variable as a training dataset. As shown in the table, Vamsa can report model and training datasets annotations for $91.88\%$ and $85.15\%$ of the scripts in the $\kaggle$ dataset. The coverage is a bit higher for the $\ntbk$ dataset.

To better understand the cases where Vamsa was not able to perform the annotation, we examined the cases where a model was not found. We identified the following reasons for the failure. (1) Some scripts called APIs commonly used for training models (e.g., \texttt{fit}), to perform other operations (e.g., feature extraction). In these cases, the KB-Based Annotator correctly did not report any model.  (2) In a few scripts, the statements used to train a model were commented out. This was not detected by our pre-processing pipeline and thus these scripts were falsely included in the final dataset. (3) Some scripts imported modules using the * notation. In these cases, Vamsa could not relate the import statement to the API calls. (4) In a few other scripts, data scientists imported a module with an alias name and used the alias when invoking the APIs. Vamsa's implementation does not currently cover such cases. We are continuously addressing these issues in our implementation.

We have also explored the cases where the KB-Based Annotator could not find a training dataset. In our analysis, we found two main reasons why. (1) In some scripts, hard-coded data (e.g., a large numpy array) was used as the training data. (2) Some APIs are not presented in our \kb and thus the annotation algorithm is not able to perform  back propagation. We note, however, that these cases could be simply covered by extending our \kb with more APIs. We further note that 
providing the ability to increase coverage by enhancing the \kb was one of the major requirements from external Vamsa users and, as such, became central to our design.

\stitle{Provenance Tracker}. Table~\ref{tbl-success-fail} shows the percentage of the cases where the Provenance Tracker can identify at least one set of features. Note that the Provenance Tracker is invoked only if the KB-Based Annotator can identify a model and its corresponding training dataset. We thus expect the coverage of this component to be bounded by the coverage of the KB-Based Annotator.

As shown in Table~\ref{tbl-success-fail}, Vamsa reports a non-empty column set for $80.85\%$ of the scripts in $\kaggle$ dataset and $74.48\%$ of the scripts in the $\ntbk$ dataset. We have also analyzed the cases that Vamsa could find both a model and a training dataset but did not discover the column set. We identified three main reasons behind this behavior:
(1) In some scripts, the columns have not been selected explicitly but based on a condition on their values (e.g., a column is in the feature set iff it contains at least $N$ non-zero values). (2) Similar to the KB-Based Annotator, some scripts required new rules to be added into the \kb for the Provenance Tracker to operate correctly. (3) Some scripts did not include any feature selection operations and thus Vamsa did not produce any output.

\begin{figure}[tb!]
\vspace{-4ex}
\begin{center}
\subfigure[Average latency breakdown.]{\label{fig-efficiency-break-down}
{\includegraphics[width=.48\columnwidth]{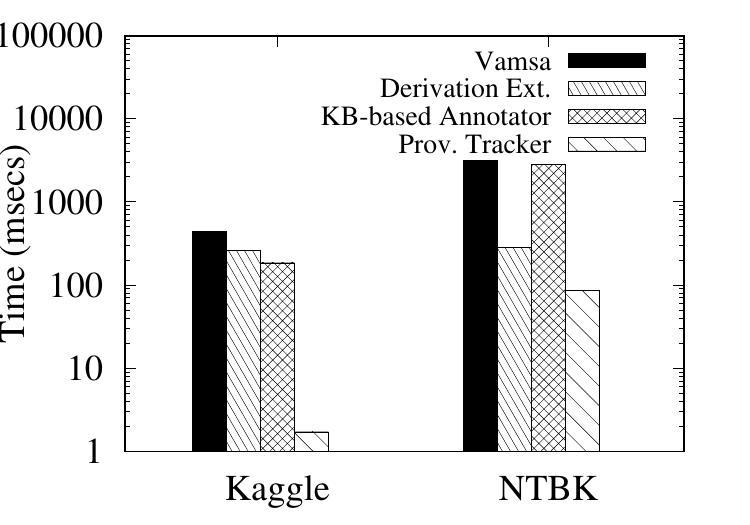}}}
\hfill
\subfigure[Average latency while varying the script size on the $\kaggle$ dataset.]{\label{fig:time-vs-loc-kaggle}
{\includegraphics[width=.48\columnwidth]{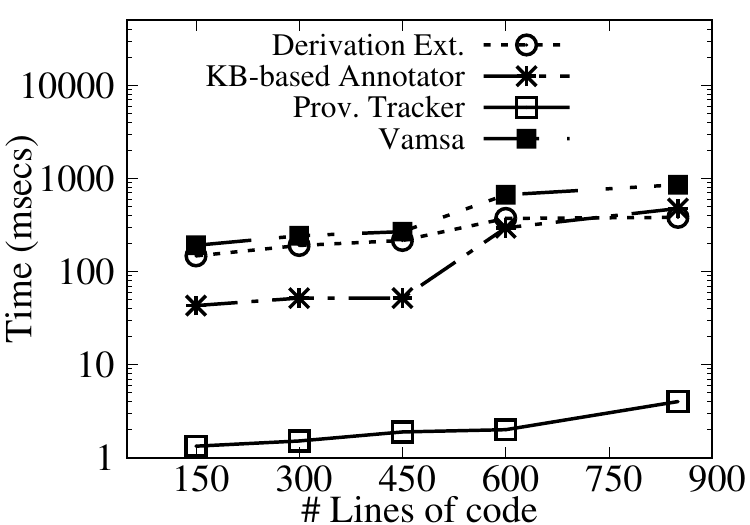}}}
\end{center}
\vspace{-4ex}
\caption{System Efficiency} \label{fig-efficiency}
\vspace{-4ex}
\end{figure}

\subsection{System Efficiency Experiments}
\label{sec:efficiency-exp}

In this set of experiments, we evaluate the latency of each component of Vamsa as well as the end-to-end latency. Note that low latency is an important requirement from Vamsa especially due to the many user scenarios where provenance needs to be extracted from many scripts in a limited amount of time (e.g., for compliance, auditing, debugging in production, or security applications).

\stitle{Breaking down the latency}. We present the individual latencies of the Derivation Extraction, KB-based Annotator, and Provenance Tracker as well as the end-to-end latency. Figure~\ref{fig-efficiency-break-down} shows the average latency results. We observe that the time spent by each component is negligible on both datasets and on average is in the order of milliseconds (ms).  One aspect that is not reflected in Figure~\ref{fig-efficiency-break-down} is the break down of the latencies of the Derivation Extractor tasks (i.e., AST generation, \LR generation, and WIR composition). For these tasks, we observed that most of the time is spent in AST generation and WIR composition. In particular, for the $\kaggle$ dataset, AST generation takes $11.1$ ms, \LR generation takes $34$ ms, and WIR composition takes $215.7$ ms. The corresponding numbers for the $\ntbk$ dataset are: $11.1$ ms, $41.1$ ms, and $227$ ms.

\stitle{Latency of Vamsa while varying the lines of code (LOCs)}. We further evaluate the latency of each component in isolation and end-to-end as the LOCs in the script vary, which is a common metric in benchmarking static analysis tools. Figure~\ref{fig:time-vs-loc-kaggle} shows the  latency of the components as the script size varies for the $\kaggle$ dataset. We see that increasing the number of LOCs in a Python script naturally increases the latency of all Vamsa components, reaching an average $0.9$s end-to-end latency ($3.9$s max) for scripts with $(600,900]$ LOCs.

\subsection{Real Use Case: Model Debugging}
We now present a real scenario where Vamsa can help automatically identify ML models that are affected by data corruption issues. Our internal Big Data platform is comprised of hundreds of thousands of machines, serving over half a million jobs daily. A large fraction of these jobs are recurring and thus have been tuned very carefully. However, some recurring jobs inevitably experience slowdowns. To understand the root cause of these slowdowns, we deploy an ML-based system that looks at various jobs parameters and runtime characteristics and identifies the cause of the slowdown~\cite{griffon-reducted}.

Developing such a system involves multiple teams: the platform team that collects and manages the  job logs, the data scientists that look into this data and develop ML models, and the team that deploys the models in production and monitors their performance. 

In one case, an engineer from the platform team identified a data corruption issue in one of the columns in the dataset.
Now, she had to identify the upstream teams that use this data and notify them on the corruption issue.  This is clearly a time-consuming and error-prone process
given that these job logs are accessed by hundreds of teams every day.
In this case, this column has been used to train the ML model responsible for root cause analysis of jobs slowdowns.

We ran Vamsa providing the script that the data scientist used to train the model and Vamsa identified the correct set of columns (included the corrupted one)
that was used to train this model. Using such a system to collect provenance information on the data scientists' scripts across the organization can help automatically notify the responsible teams on data corruption issues
 so that they can quickly take the appropriate steps to fix their models.

\section{Related Work}
\label{sec:related}

We describe relevant related work from three areas:

\stitle{Model management systems}. There has been an emerging interest in systems that manage the lifecycle of ML models~\cite{schelter2018challenges, garcia2018context, vartak2016m, miao2017towards, schelter2017automatically, miao2018provdb,vartak2018mistique}. ModelDB~\cite{vartak2016m} stores trained models to enable querying of metadata and artifacts by exposing a logging API for a specific set of libraries. ModelHub~\cite{miao2017towards} is a fine-grained versioning system for ML artifacts with a focus on deep learning. Amazon's ML experiments system~\cite{schelter2017automatically} tracks provenance of ML experimentation data. This system automated the provenance extraction for SparkML~\cite{meng2016mllib} and scikit-learn~\cite{pedregosa2011scikit} pipelines whenever a logical abstraction of operations is available. ProvDB~\cite{miao2018provdb} focuses on efficiently storing and querying ML provenance data. 
In contrast to these systems, Vamsa focuses on the tracking task and, as such, is complementary to systems that focus on other management tasks (e.g., storing or querying). Furthermore, Vamsa introduces design principles (i.e., does not require developers to modify their code, operates on top of any library, and tracks provenance at static analysis time) that are important for end-users, yet not provided by prior systems.

\stitle{Provenance in databases}. Capturing provenance on top of SQL queries is an extensively studied area~\cite{ikeda2009data, cheney2009provenance} driven by an immense amount of applications~\cite{namaki2019answering,scorpion,pnl:2017:deutch, psallidas2018smoke,psallidas2018hilda,ragan2016characterizing,dbnotes2005chiticariu,cui2000linagetrace,gdpr}. %
In contrast to this line of work, Vamsa is designed to capture provenance in data science scripts as opposed to SQL queries. As such, Vamsa introduces provenance capture techniques tailored to the semantics of imperatively specified data science logic (in Python). Furthermore, Vamsa treats models and datasets as first-class citizens of the captured provenance information, all the while exposing (and semantically annotating) data flows of data science logic. This allow us to better drive applications in the data science space.

\stitle{Workflow management systems}. %
Workflow management systems collect and manage the provenance information to enable experiment sharing~\cite{freire2007provenance}. Closer to our work are the Starflow~\cite{angelino2010starflow, angelino2011provenance}, noWorkflow~\cite{pimentel2017noworkflow}, and YesWorkflow~\cite{mcphillips2015yesworkflow} systems. StarFlow statically analyzes a Python program to build provenance traces at the level of functions. noWorkflow captures various forms of provenance information by analyzing scripts during execution. Starflow and YesWorkflow require modifications to the users' script, while noWorkflow handles unmodified programs.  Our work differs from this line of work in a vein similar to the other two lines of work discussed above. We note, however, that an interesting direction is to extend our techniques to account for runtime information.

\section{Conclusions and Future Work}
\label{sec:conclusion}

In this paper, we introduced the problem of ML provenance tracking---a fundamental type of provenance information that enables multiple applications including model debugging; compliance; and model maintenance. Our evaluation shows that it is indeed possible to recover this type of provenance with a very high precision and recall across large corpora of Python scripts.

There are many areas to further explore. First, incorporating runtime information can be useful for cases that are undecidable under our static analysis setting. Second, identifying finer-grained provenance information between data sources and ML models (e.g., partitions of a data source were used for training) can better assist upstream applications. Finally, automatically populating the knowledge base is also an important direction for future work.

\balance

\bibliographystyle{abbrv}
\bibliography{paper}

\newpage
\section*{Reproducibility}

In this section we report additional information regarding our experimental evaluation of Section~\ref{sec:exp}.
We are taking the following steps to ensure reproducibility of our experiments:

\textbf{Datasets}: As mentioned in Section~\ref{sec:exp}, we used a large corpus of \textbf{publicly} available Python scripts ($26K$).
This dataset is available at~\cite{vamsaurl}.

\textbf{Experiments with Labeled Datasets}: In Section 7.2, we presented experiments with scripts for which we have manually extracted the provenance information (ground truth).
The ground truth for the scripts that were used in these experiments is also available at~\cite{vamsaurl}.

\textbf{Algorithms}: The pseudocode for all the algorithms used by Vamsa is either presented in this paper, or is available in our technical report~\cite{tr}. 

\textbf{Knowledge Base:} A file with the data stored in the knowledge base that we used for our experiments is available at~\cite{vamsaurl}.

\begin{figure}[bh]
\centering
\centerline{\includegraphics[scale=0.65]{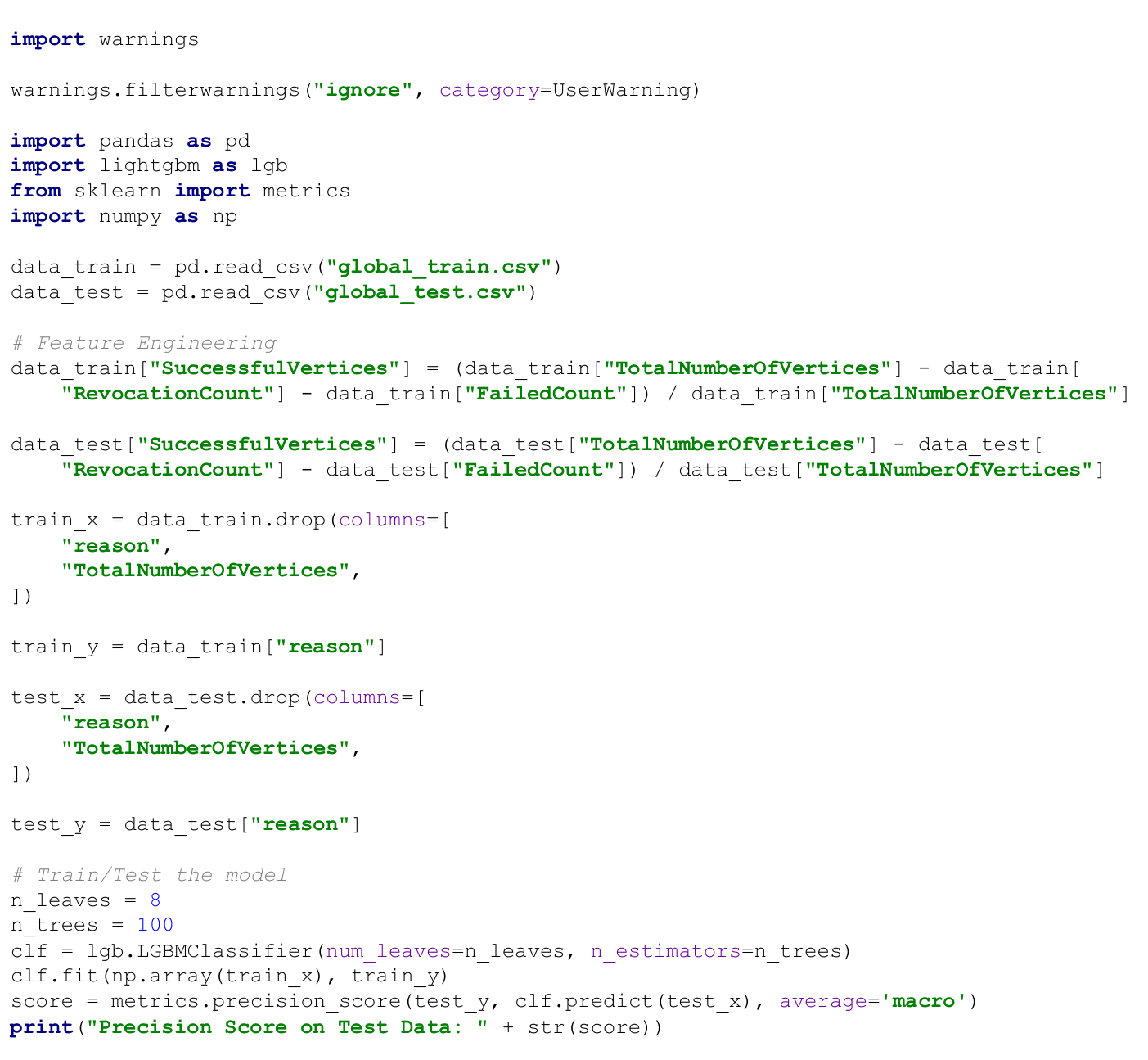}}
 \caption{{\small Simplified data science script used in our real example}}
\label{fig-real-example}
\end{figure}

\textbf{Real Use Case}: In section 7.5, we presented a real scenario encountered in production that highlights the significance of using tools such as Vamsa to collect provenance information from ML scripts.
Figure~\ref{fig-real-example} presents a simplified but representative version of the script that is used in production.

\end{document}